\documentclass{bmvc2k}

\usepackage{times}
\usepackage{epsfig}
\usepackage{graphicx}
\usepackage{amsmath}
\usepackage{amssymb}
\usepackage{algorithm}
\usepackage[noend]{algpseudocode}
\usepackage{caption}
\usepackage{array,booktabs}
\usepackage{siunitx}
\usepackage[dvipsnames]{xcolor}
\usepackage{booktabs}

\title{Mixup-CAM: Weakly-supervised Semantic Segmentation via Uncertainty Regularization}

\addauthor{Yu-Ting Chang}{}{1}
\addauthor{Qiaosong Wang}{}{2}
\addauthor{Wei-Chih Hung}{}{1}
\addauthor{Robinson Piramuthu}{}{2}
\addauthor{Yi-Hsuan Tsai}{}{3}
\addauthor{Ming-Hsuan Yang}{}{14}
\addinstitution{
 University of California, Merced
}
\addinstitution{
 eBay Inc.
}
\addinstitution{
 NEC Labs America
}
\addinstitution{
 Google Research
}
\runninghead{Chang \lowercase{et al.}}{Mixup-CAM: Weakly-supervised Semantic Segmentation}

\begin{document}

\maketitle

\begin{abstract}

Obtaining object response maps is one important step to achieve weakly-supervised semantic segmentation using image-level labels.
However, existing methods rely on the classification task, which could result in a response map only attending on discriminative object regions as the network does not need to see the entire object for optimizing the classification loss.
To tackle this issue, we propose a principled and end-to-end trainable framework to allow the network paying attention to other parts of the object, while producing a more complete and uniform response map.
Specifically, we introduce the mixup data augmentation scheme into the classification network and design two uncertainty regularization terms to better interact with the mixup strategy.
In experiments, we conduct extensive analysis to demonstrate the proposed method and show favorable performance against state-of-the-art approaches.

\end{abstract}

\section{Introduction}
Semantic segmentation is one of the fundamental tasks in computer vision, with a wide range of applications such as image editing and scene understanding.
In order to obtain reliable models and achieve promising performance, recently deep neural network (DNN) based methods \cite{fcn_pami,deeplab,dilated} are learned from fully-supervised data that requires pixel-wise semantic annotations.
However, acquiring such pixel-wise annotations is usually time-consuming and labor-intensive, which limits the application potentials in the real world.
As a result, numerous approaches tackle this issue via training models only on weakly-annotated data, \textit{e.g.}, image-level \cite{ahn2018learning,kolesnikov2016seed,pathak2015constrained,pinheiro2015image}, bounding box \cite{papandreou2015weakly,dai2015boxsup,khoreva_CVPR17}, point-level \cite{Bearman_ECCV16}, scribble-based \cite{lin2016scribblesup,Vernaza_CVPR17}, or video-level \cite{Chen_IJCV_2020,Zhong_ACCV_2016,Tsai_ECCV_2016} labels.
In this paper, we focus on utilizing the image-level label, which is the most efficient scheme for weak annotations but also a challenging scenario.

Existing weakly-supervised semantic segmentation (WSSS) algorithms mainly operate with three steps: 1) localizing objects via a categorical response map, 2) refining the response map to generate pseudo annotations, and 3) training the semantic segmentation network using pseudo ground truths.
Recent methods \cite{ahn2018learning,huang2018weakly,wang2018weakly,wei2018revisiting} have achieved significant progress for WSSS, but most of them focus on improving the latter two steps.
Since the success of these sequential steps hinges on the quality of the initial response map generated in the first step, in this paper we present an effective solution to localize objects.

\begin{figure*}[t]
	\centering
	\includegraphics[width=0.95\linewidth]{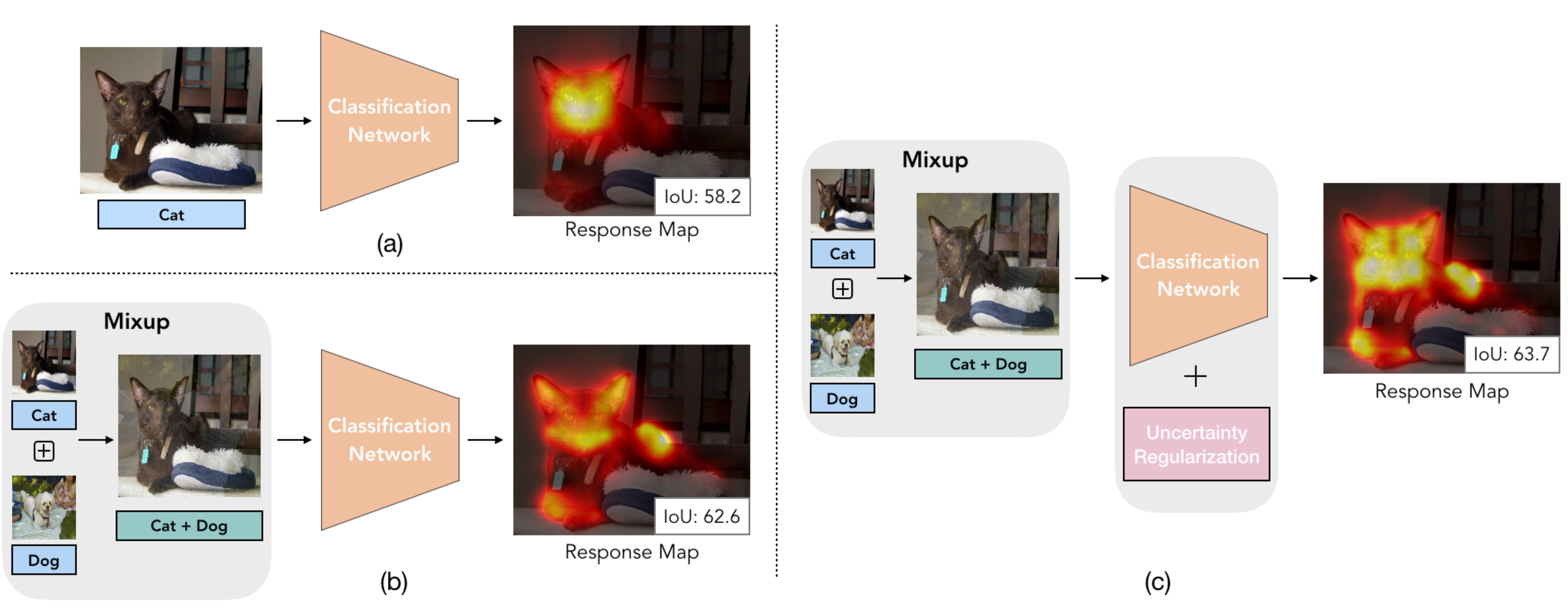}\\
	\vspace{3mm}
	\caption{Comparisons of (a) the original CAM method; (b) CAM + the mixup data augmentation; and (c) the proposed Mixup-CAM framework that integrates the mixup scheme and the uncertainty regularization.
	Compared to (a) and (b), our final response map (c) attends to other object parts with more uniformly distributed response.
	}
	\label{fig: teaser}
	\vspace{-5mm}
\end{figure*}

One common practice to produce the initial response map is using class activation map (CAM) \cite{zhou2016learning}. However, since CAM is typically supervised by a classification loss that could be sufficiently optimized through seeing only a small portion of objects, the generated response map usually only attends on partial regions (see Figure \ref{fig: teaser}(a)).
To tackle this issue, recent methods \cite{lee2019ficklenet,jiang2019integral} make efforts to improve the response map via using the dropout strategy that increases the model uncertainty or aggregating maps produced at different stages to see more object parts.
However, there remains a challenge whether there are better loss function designed to explicitly facilitate the model training and produce better response maps, which are not addressed in prior works.

In this paper, we propose a principled and end-to-end trainable network with loss functions designed to systematically control the generation of the response map.
First, inspired by the mixup data augmentation in \cite{zhang2017mixup}, we observe that including mixup could effectively calibrate the model uncertainty on overconfident predictions \cite{Thulasidasan2019mixup} and in return enables the model to attend to more object regions.
However, it is challenging to control the mixup augmentation process and the model uncertainty, due to non-uniform response distributions (see Figure \ref{fig: teaser}(b)), which may affect subsequent response refinement steps.
Therefore, we introduce another two loss terms to the mixup process by regularizing the class-wise uncertainty and the spatial response distribution.
We refer to our model as \textit{Mixup-CAM} and show that the produced response map is more complete and balanced across object regions (see Figure \ref{fig: teaser}(c)), which facilitates the latter response refinement and segmentation model training steps.

We conduct quantitative and qualitative experiments to demonstrate the effectiveness of the proposed Mixup-CAM method on the PASCAL VOC 2012 dataset \cite{PASCAL_VOC_2010_Data}.
To the best of our knowledge, our algorithm is the first to demonstrate that mixup could improve the WSSS task on complicated multi-labeled images, along with other designed loss functions to produce better response maps.
In addition, we present the ablation study and more analysis to validate the importance of each designed loss.
Finally, we show that our method achieves state-of-the-art semantic segmentation performance against existing approaches.

\section{Related Work}

\paragraph{Initial Prediction for WSSS.}
Initial cues are essential for segmentation tasks since they provide reliable priors to generate segmentation maps. The class activation map (CAM)  \cite{zhou2016learning} is a common practice for localizing objects. It highlights class-specific regions that are served as the initial cues. However, since the CAM model is trained by a classification network, it tends to attend to small discriminative parts of the object, leading to incomplete initial masks.
Several methods have been developed to alleviate this problem. Approaches like \cite{singh2017hide, wei2017object, zhang2018adversarial, li2018tell} deliberately hide or erase the regions of an object, forcing the model to look for more diverse parts. However, such strategies require iterative model training and response aggregation steps. After gradual expansion of the attention regions, non-object regions are prone to be activated, which leads to inaccurate attention maps. Other algorithms \cite{hou2018self, zeng2019joint} use both object and background cues to prevent the attention map from including more background regions, yet pixel-level saliency labels are used.

Instead of using the erasing scheme, recently the FickleNet approach \cite{lee2019ficklenet} introduces the stochastic feature selection to obtain a diverse combination of locations on feature maps. Moreover, the OAA method \cite{jiang2019integral} adopts an online attention accumulation strategy to collect various object parts discovered at different training stages. By aggregating the attention maps, it could obtain an initial cue that contains a larger region of the object. 
Unlike methods that mitigate the problem by discovering complementary regions via iterative erasing steps or consolidating attention maps, our proposed approach aims at harnessing the uncertainty in end-to-end classification learning. In addition, by regularizing both class-wise uncertainties and spatial response distributions, our approach averts the attention from focusing on small parts of the semantic objects, hence producing much improved response maps.

\vspace{-3mm}
\paragraph{Response Refinement for WSSS.}
Various approaches \cite{ahn2018learning,fan2018cian, fan2018associating,huang2018weakly,kolesnikov2016seed,wang2018weakly,wei2018revisiting} are proposed to refine the initial cue by expanding the region of attention maps. Other methods \cite{ahn2018learning, fan2018cian, fan2018associating} are developed using affinity learning. The recent SSDD scheme \cite{shimoda2019self} proposes a difference detection module to estimate and reduce the gap between the initial mask and final segmentation results in a self-supervised manner.
However, the performance of these methods is limited as initial seeds are still obtained from CAM-like methods. If these seeds only come from the discriminative parts of the object, it is difficult to expand regions into non-discriminative parts. Moreover, the initial prediction may produce wrong attention regions, which would lead to even more inaccurate regions in subsequent refinement steps.

\vspace{-3mm}
\paragraph{Label-preserving \emph{vs.} Non-preserving Augmentations.}
Data augmentation is a common regularization technique in both supervised and unsupervised learning \cite{cubuk2019autoaugment,lim2019fast,he2019momentum}. Conventional data augmentation techniques such as scaling, rotation, cropping, color augmentation, and Gaussian noise can change the pixel values of an image without altering its labels. These label-preserving transformations are commonly applied in training deep neural networks to improve the model generalization capabilities.

Recent work has demonstrated that even non-label-preserving data augmentation can be surprisingly effective. Explicit label smoothing has been adopted successfully to improve the performance of deep neural models. The Mixup method \cite{zhang2017mixup} is proposed to train a neural network on a convex combination space of image pairs and their corresponding labels. It has been proven effective for the classification task and increases the robustness of neural networks. Numerous Mixup variants \cite{berthelot2019mixmatch, summers2019improved, Yun2019mixup, verma2018manifold, Thulasidasan2019mixup,guo2019mixup} have been proposed to extend mixup for better prediction of uncertainty and calibration of the DNNs. These methods exhibit shared similarity of producing better-generalized models.

\vspace{-3mm}
\paragraph{Entropy Regularization.}
Aside from mixup schemes for predictive uncertainty, another common uncertainty measure is entropy, which could act as a strong regularizer in both supervised and semi-supervised learning \cite{pereyra2017regularizing,grandvalet2005semi}.
In particular, \cite{pereyra2017regularizing} discourages the neural network from being over-confident by penalizing low-entropy distributions, while \cite{grandvalet2005semi} utilizes entropy minimization in a semi-supervised setting as a training signal on unlabeled data.
In this paper, we also adopt the entropy-based loss to regularize the uncertainty, coupled with the mixup data augmentation for producing better response maps on objects.

%
%

\begin{figure*}[t]
	\centering
	\includegraphics[width=0.95\linewidth]{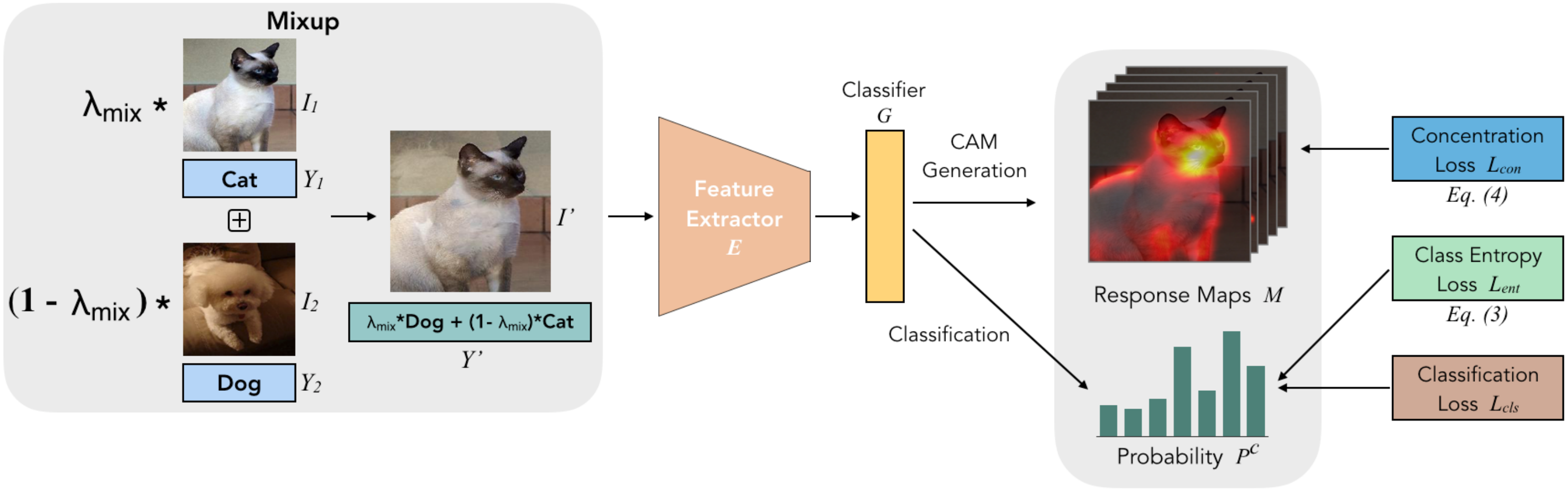}\\
	\vspace{3mm}
	\caption{Overview of Mixup-CAM. We perform mixup data augmentation on input images with their corresponding labels via \eqref{eq:mixup} and pass the mixed image through the feature extractor $E$ and the classifier $G$ to obtain the probability score $P^c$ for each category $c$. For loss functions, in addition to the classification loss $\mathcal{L}_{cls}$ on mixup samples, we design two terms to regularize class-wise entropy ($\mathcal{L}_{ent}$ via \eqref{eq:l_ent}) and spatial distribution on CAM ($\mathcal{L}_{con}$ via \eqref{eq:l_con}).
	}
	\label{fig: framework}
	\vspace{-5mm}
\end{figure*}

\section{Weakly-supervised Semantic Segmentation}
We first describe the overall algorithm and introduce details of the proposed Mixup-CAM framework with loss functions designed to improve the initial response map. 
We then detail how to generate the final semantic segmentation results.

\subsection{Algorithm Overview}
One typical way to generate response maps for annotated object categories is to use
CAM \cite{zhou2016learning}. 
However, these response maps tend to focus on discriminative object parts, which are less effective for the WSSS task.
One reason is that CAM relies on the classification loss, which only requires partial object regions to be activated during training. As a result, when the objective is already optimized with high confidence, the model may not attempt to learn other object parts.

In this paper, we propose to integrate the idea of mixup data augmentation \cite{zhang2017mixup}, thereby calibrating the uncertainty in prediction \cite{Thulasidasan2019mixup} as well as allowing the model to attend to other regions of the image.
Although we find that adding mixup could improve the response map, sometimes the response could diverge too much, resulting in more false-positive object regions.
To further regularize this uncertainty, we introduce two additional loss terms: the spatial loss and the class-wise loss.
We illustrate the overall model and loss designs in Figure~\ref{fig: framework} and provide more details in the next subsection.

After receiving the initial response map, we utilize the method in \cite{ahn2018learning} to expand and refine the response.
Finally, we generate pseudo ground truths from the refined response and train a semantic segmentation network to obtain the final segmentation output.
Note that while we focus on the first step of the initial response map in this paper, the succeeding two steps could be replaced with alternative modules or models.

\subsection{Mixup-CAM}
\paragraph{CAM Generation.}
We first describe the CAM method for producing the initial response map as our baseline (see Figure \ref{fig: teaser}(a)). The base network begins with a feature extractor $E$, followed by a global average pooling (GAP) layer and a fully-connected layer $G$ as the output classifier.
Next, given an input image $I$ with its image-level labels $Y$, the network is trained with a multi-label classification loss $\mathcal{L}_{cls}(Y, G(E(I)))$ following \cite{zhou2016learning}.
After training this classification network, the activation map $M^c$ for each category $c$ is obtained by applying the $c$-channel classifier weight $\theta_G^c$ on the feature map $f = E(I)$:
\begin{equation}
    M^c = \theta_G^{^c \top} f.
    \label{eq:cam}
\end{equation}
%
Finally, the response is normalized by the maximum value of $M^c$.

\vspace{-3mm}
\paragraph{Mixup Data Augmentation.}
Since the original classification network could easily obtain high confidence, in which the generated CAM only attends to small discriminative object parts, we utilize mixup data augmentation to calibrate the uncertainty in prediction \cite{Thulasidasan2019mixup}.
Given an image pair $\{I_1, I_2\}$ and its label $\{Y_1, Y_2\}$ randomly sampled from the training set, we augment an image $I'$ and its label $Y'$ via:
\begin{align}
    I' & = \lambda_{mix} I_1 + (1-\lambda_{mix}) I_2, \notag \\
    Y' & = \lambda_{mix} Y_1 + (1-\lambda_{mix}) Y_2,
    \label{eq:mixup}
\end{align}
where $\lambda_{mix}$ is sampled from the \textit{Beta}$(\alpha, \alpha)$ distribution following \cite{zhang2017mixup}.
Using this augmented data, we feed it into the classification network to minimize the loss $\mathcal{L}_{cls}(Y', G(E(I')))$ and follow the same procedure in \eqref{eq:cam} to produce the response map (see Figure \ref{fig: teaser}(b)).

Compared to the original CAM generation, our network no longer receives a pure image but a mixed image that could have multiple objects with their weights based on $\lambda_{mix}$ as in \eqref{eq:mixup}.
Therefore, the predictive uncertainty could be enhanced, leading to smoother output distributions and enforcing the model to pay attention to other regions in the image in order to satisfy the classification loss $\mathcal{L}_{cls}(Y', G(E(I')))$.
\vspace{-3mm}
\paragraph{Uncertainty Regularization.}
Although mixup could improve the response map by looking at other parts in the image, sometimes the response could become too divergent and thus attend to pixels non-uniformly, \textit{e.g.}, Figure \ref{fig: teaser}(b).
This is attributed to the difficulty in controlling the quality of mixed images, especially when the model faces more complicated images such as PASCAL VOC, \textit{e.g.}, an object could appear at various locations of the image with noisy background clutters.
%

%
To further facilitate the mixup process, we propose to self-regularize the uncertainty via class-wise loss and spatial loss terms.
The first term is to directly minimize the entropy in output prediction from the classifier to reduce uncertainty:
\begin{equation}
    \mathcal{L}_{ent}(G(E(I'))) = - \frac{1}{HW} \sum_{h,w} \sum_{c \in C} P^c(h,w) \log P^c(h,w),
    \label{eq:l_ent}
\end{equation}
where $C$ is the category number and $P^c \in \mathbb{R}^{H \times W}$ is the output probability for category $c$.
Since our classifier $G$ outputs multi-label probability, we concatenate the probabilities and normalize them by the maximum value, then calculate the final $P$ with \textit{softmax}.

Although the first term has the ability to minimize the uncertainty, it does not explicitly operate on the response map. To better balance the distribution on the response, we utilize a concentration loss similar to \cite{hung2019scops} and apply it directly on CAM for each category (\textit{i.e.}, $M^c$), which encourages activated pixels to be spatially close to the response center:
\begin{equation}
    \mathcal{L}_{con}(M) = \sum_{c \in \bar{C}} \sum_{h,w} ||\langle h,w\rangle - \langle \mu_h^c,\mu_w^c\rangle||^2 \cdot \hat{M}^c(h,w),
    \label{eq:l_con}
\end{equation}
where $\mu_h^c = \sum_{h,w} h \cdot \hat{M}^c(h,w)$ is the center in height for category $c$ (similarly for $\mu_w^c$), $\hat{M}^c$ is the normalized response of $M^c$ to represent a spatially distributed probability map.
Note that, here we only calculate the concentration loss on presented categories $\bar{C}$ as provided in the image-label $Y$ to avoid confusing the model with invalid categories.

\vspace{-3mm}
\paragraph{Overall Objective.}
We have described our proposed Mixup-CAM framework, including mixup data augmentation in \eqref{eq:mixup} and two regularization terms, \textit{i.e.}, \eqref{eq:l_ent} and \eqref{eq:l_con}. To train the entire model in an end-to-end fashion, we perform the online mixup procedure and jointly optimize the following loss functions:
\begin{equation}
    \mathcal{L}_{all} = \mathcal{L}_{cls}(I', Y') + \lambda_{ent} \mathcal{L}_{ent}(I') + \lambda_{con} \mathcal{L}_{con}(M).
    \label{eq:l_all}
\end{equation}
For simplicity, we omit the detailed notation inside each loss term. We also note that $M$ is produced online via computing \eqref{eq:cam} on valid categories in each forward iteration.

\subsection{Implementation Details}
\paragraph{Classification Network.}
Similar to \cite{ahn2018learning}, 
we use the ResNet-38 architecture \cite{wu2019wider} as our classification network, which consists of 38 convolution layers with wide channels, followed by a $3 \times 3$ convolution layer with 512 channels for better adaptation to the classification task, a global average pooling layer, and two fully-connected layers for classification.
In training, we adopt the pre-trained model on ImageNet \cite{deng2009imagenet} and finetune it on the PASCAL VOC 2012 dataset.
Typical label-preserving data augmentations, \textit{i.e.}, horizontal flip, random cropping, random scaling, and color jittering, are utilized on the training set.

We implement the proposed Mixup-CAM framework using PyTorch with a single Titan X GPU with 12 GB memory. 
For training the classification network, we use the Adam optimizer \cite{kingma2014adam} with initial learning rate of 1e-3 and the weight decay of 5e-4.
For mixup, we use $\alpha = 0.2$ in the \textit{Beta}$(\alpha,\alpha)$ distribution.
For uncertainty regularization, we set $\lambda_{ent}$ as 0.02 and $\lambda_{con}$ as 2e-4. Unless specified otherwise, we use the same parameters in all the experiments.
In the experimental section, we show studies for the sensitivity of different parameters.
\vspace{-3mm}
\paragraph{Semantic Segmentation Generation.}

Based on the response map generated by our Mixup-CAM, we adopt the random walk approach via affinity \cite{ahn2018learning} to refine the response and produce pixel-wise pseudo ground truths for semantic segmentation.
In addition, similar to existing methods, we adopt dense conditional random fields (CRF) \cite{crf} to further refine the response and obtain better object boundaries.
Finally, we utilize the Deeplab-v2 framework \cite{deeplab} with the ResNet-101 architecture \cite{He_2016_CVPR} and train the segmentation network

\vspace{-2mm}
\section{Experimental Results}
%
In this section, we present our main results of the proposed Mixup-CAM method for the WSSS task.
First, we show that our approach achieves better initial response maps and further improves the subsequent refinement step.
Second, we demonstrate the importance of each designed component.
Finally, we provide evaluations on final semantic segmentation outputs in the PASCAL VOC dataset \cite{PASCAL_VOC_2010_Data} against the state-of-the-art approaches.
More results can be found in the supplementary material. We will make our code and models available to the public.

\vspace{-2mm}
\subsection{Evaluated Dataset and Metric}
We conduct experiments on the PASCAL VOC 2012 semantic segmentation benchmark \cite{PASCAL_VOC_2010_Data} with 21 categories, including one background class. 
%
Following existing WSSS methods, we use augmented 10,528 training images \cite{hariharan2011semantic} to train our network. 
For evaluation of response maps of the training set, we use the set without augmentation with 1,464 examples, following the setting in \cite{ahn2018learning}.
For final semantic segmentation results, we use 1,449 images in the validation set to compare our results with other methods\footnote{Although there is a test set that can be evaluated on the official PASCAL VOC website, by the submission deadline the website is still out of service for returning the evaluated performance.}. 
In all experiments, the mean Intersection-over-Union (mIoU) ratio is used as the evaluation metric. 
%

\begin{figure*}[t]
	\centering
	\includegraphics[width=0.95\linewidth]{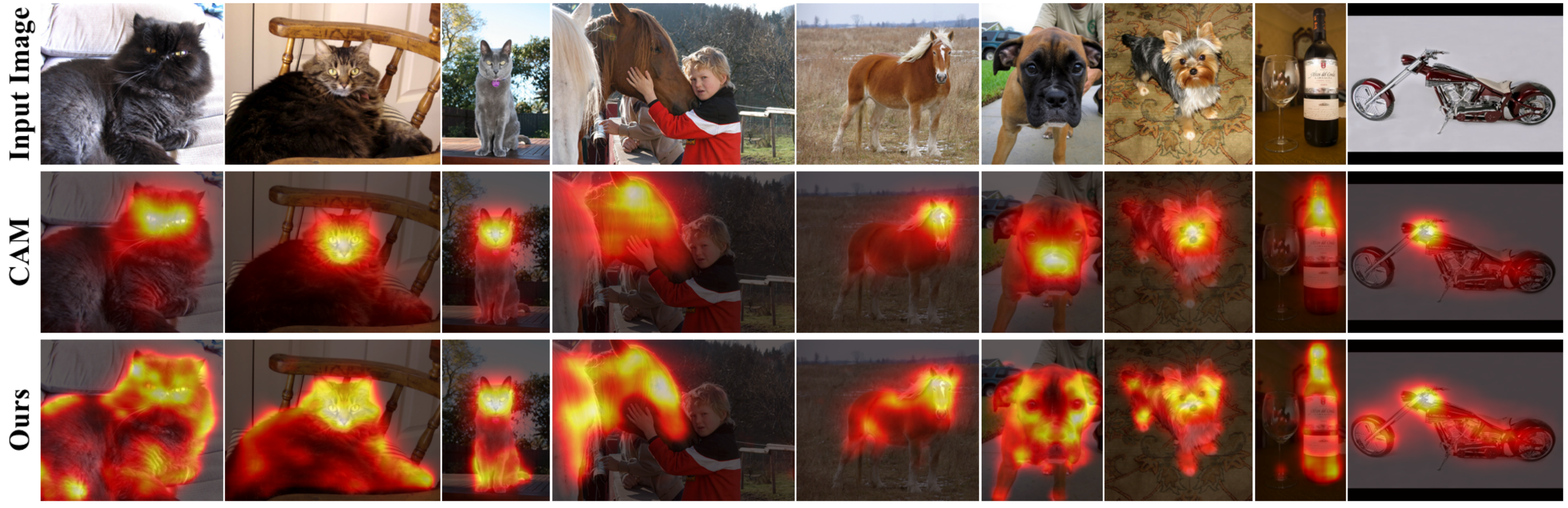}
	\vspace{3mm}
	\caption{Sample results of initial responses. Our approach often produces the response map that covers more complete region of the object (i.e., attention on the body of the animal), while the initial cue obtained by CAM \cite{zhou2016learning} is prone to focus on small discriminative parts.}
	\label{fig: response_map}
	\vspace{-2mm}
\end{figure*}

\begin{table}[!t]
	\caption{IoU results of CAM and its refinement on the PASCAL VOC training set.}
	\vspace{3mm}
	\small
	\centering
	\renewcommand{\arraystretch}{1.1}
	\setlength{\tabcolsep}{6pt}
	\begin{tabular}{lcc}
		\toprule
		Method & CAM & CAM + Refinement \\
		\midrule
	
		AffinityNet \cite{ahn2018learning} & 48.0 & 58.1 \\
		Mixup $\mathcal{L}_{cls}$ & 49.3 & 60.5 \\
		Mixup $\mathcal{L}_{cls}$ + $\mathcal{L}_{ent}$ & 49.5 & 61.6 \\
		Mixup $\mathcal{L}_{cls}$ + $\mathcal{L}_{con}$ & 49.9 & 61.7 \\
		Mixup $\mathcal{L}_{cls}$ + $\mathcal{L}_{ent}$ + $\mathcal{L}_{con}$ & \textbf{50.1} & \textbf{61.9} \\
		
		\bottomrule
	\label{table: ablation_loss}
	\end{tabular}
	\vspace{-7mm}
\end{table}

\vspace{-2mm}
\subsection{Ablation Study and Analysis}

\paragraph{Improvement on Response Map.}

We first present results of the initial and refined response maps.
In Table \ref{table: ablation_loss}, we show the performance for the original CAM used by the baseline AffinityNet \cite{ahn2018learning}, our CAM using the mixup data augmentation (Mixup $\mathcal{L}_{cls}$), and our final Mixup-CAM with mixup and uncertainty regularization (Mixup $\mathcal{L}_{cls}$ + $\mathcal{L}_{ent}$ + $\mathcal{L}_{con}$).
In both results of CAM and its refinement, our IoU improvements are consistent after gradually adding the mixup augmentation and regularization.
In addition, Figure \ref{fig: response_map} shows some example results of the initial response, which illustrates that our Mixup-CAM is able to make the network attend to more object parts and produce more uniform response distributions on objects.


\begin{figure*}[t]
	\centering
	\includegraphics[width=0.95\linewidth]{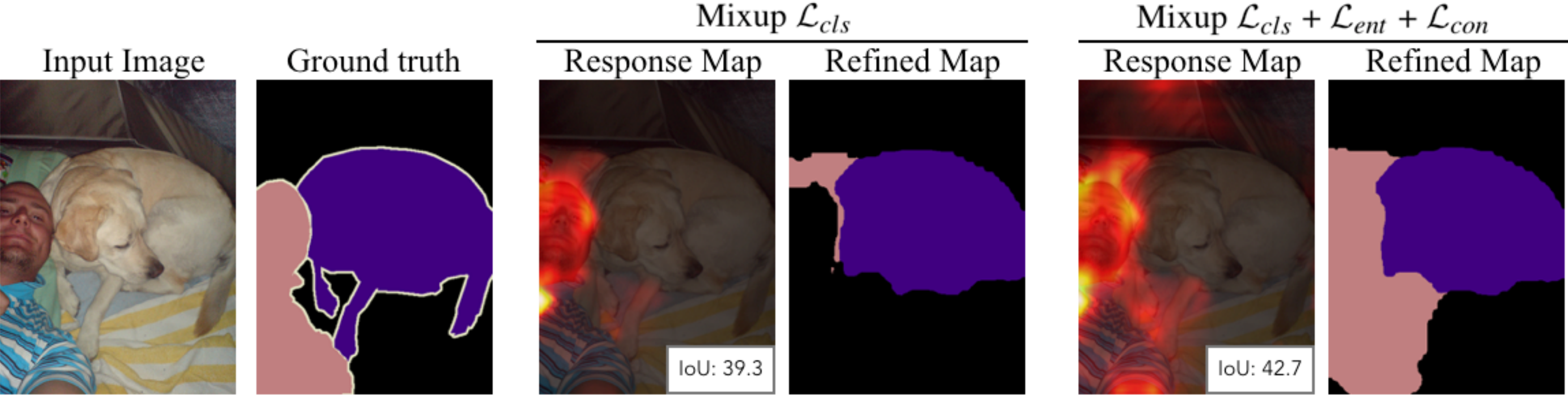}\\
	\vspace{3mm}
	\caption{Enhancement on refinement. Our regularization enforces a more uniform response on objects, which can facilitate the refinement step. The examples illustrate that the IoU difference of the resultant refined map is significantly larger than the one of initial response.}
	\label{fig: rw_diff}
	\vspace{-3mm}
\end{figure*}

\begin{figure*}[t]
	\centering
	\includegraphics[width=0.95\linewidth]{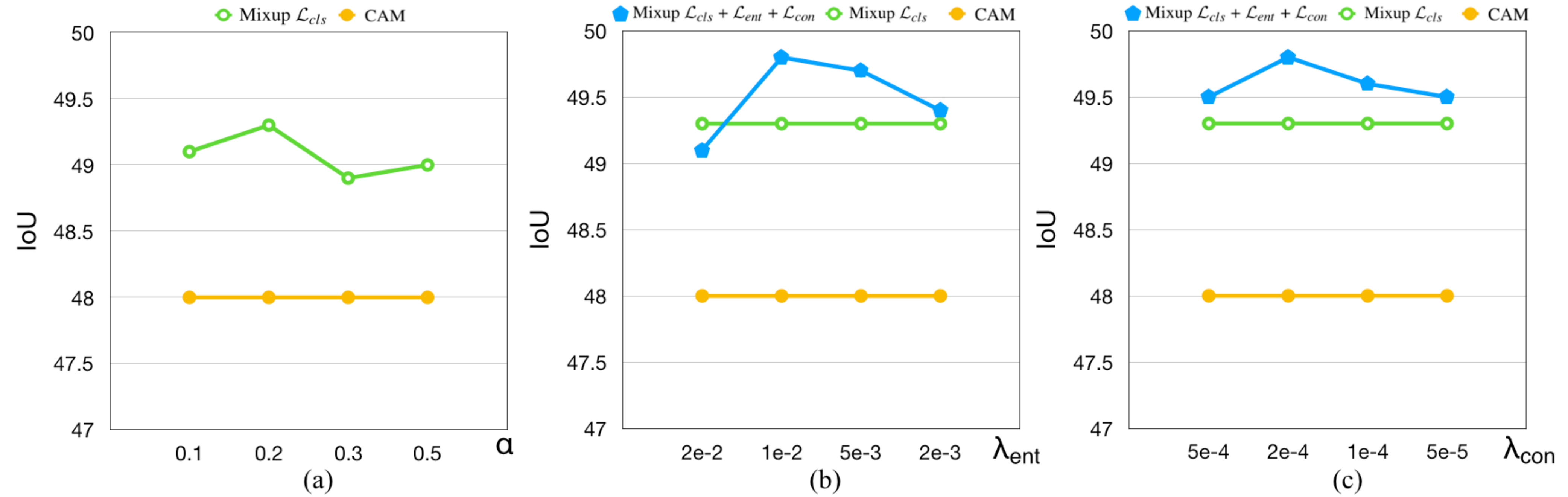}
	\vspace{3mm}
	\caption{Sensitivity analysis for parameters. (a) $\alpha$ for mixup augmentation;
	(b) $\lambda_{ent}$ and (c) $\lambda_{con}$ for uncertainty regularization.
	}
	\label{fig: para}
	\vspace{-5mm}
\end{figure*}

\vspace{-3mm}
\paragraph{Effectiveness of Regularization.}

One interesting aspect we find is that adding regularization could enhance the effectiveness of the refinement step.
In Table \ref{table: ablation_loss}, compared to Mixup $\mathcal{L}_{cls}$, adding either $\mathcal{L}_{ent}$ (3rd row) or $\mathcal{L}_{con}$ (4th row) improves the CAM IoU by 0.2\% and 0.6\% respectively.
Nevertheless, with the refinement, the corresponding improvements in IoU is 1.1\% and 1.2\%, which are larger than the ones before refining the response.
This is because our regularization enforces a more uniform response on objects, which greatly facilitates the refinement step (\textit{e.g.}, via region expanding).
In addition, we illustrate one example in Figure \ref{fig: rw_diff}, where the IoU difference of initial response is relatively small, but the resultant refined map could differ significantly.


\vspace{-3mm}
\paragraph{Parameter Sensitivity.}
In this paper, we mainly study three parameters in our Mixup-CAM framework, i.e., $\alpha$ for mixup regularization and $\{\lambda_{ent}, \lambda_{con}\}$ in uncertainty regularization.
In Figure \ref{fig: para}(a), when increasing the $\alpha$ value, the \textit{Beta} distribution would become more uniform, which encourages a more uniform $\lambda_{mix}$ in \eqref{eq:mixup} and results in mixed images that are more challenging to optimize.
Nevertheless, the IoUs of Mixup $\mathcal{L}_{cls}$ under various $\alpha$ are consistently better than the CAM baseline.
For regularization terms, we fix $\lambda_{con} = 2e-4$ and adjust $\lambda_{ent}$ in Figure \ref{fig: para}(b), while fixing $\lambda_{ent} = 0.2$ and change $\lambda_{con}$ in Figure \ref{fig: para}(c).
Both figures show that these two parameters are robust to the performance under a wide range.


\subsection{Semantic Segmentation Performance}
After generating the pseudo ground truths using the refined response map, we use them to train the semantic segmentation network.
First, we compare our method with state-of-the-art algorithms using the ResNet-101 architecture or other similarly powerful ones in Table \ref{table: compare_sota}.
%
%
Note that, while our method focuses on improving the initial responses on the object, most methods aim to improve the refinement step or segmentation network training.
In Table \ref{table: compare_sota_detail}, we further present detailed performance for each category.
We show two groups of results without (top rows) or with (bottom rows) applying CRF \cite{crf} to refine final segmentation outputs.
Compared to the recent FickleNet \cite{lee2019ficklenet} approach that also tries to improve the initial response map, our proposed Mixup-CAM shows favorable performance in final semantic segmentation results.
%
%
%



\begin{table}[!t]
\caption{Comparison of WSSS methods using image-level labels on the PASCAL VOC 2012 validation set. $\checkmark$ indicates the methods that focus on improving the initial response. The result of $\dagger$ on AffinityNet is re-produced by training the same ResNet-101 as our pipeline.}
\vspace{3mm}
\label{tab:orbital_data}
\centering
\scriptsize
\renewcommand{\arraystretch}{1.1}
\setlength{\tabcolsep}{4pt}
\begin{tabular}{lccc} 
\toprule
\text{Method} & {Backbone} & {Init. Resp.} & {IoU on Val} \\
\midrule
    MCOF \textsubscript{CVPR'18} \cite{wang2018weakly} & ResNet-101 &  & 60.3 \\

    DCSP \textsubscript{BMVC'17} \cite{chaudhry2017discovering} & ResNet-101 &  & 60.8 \\

    DSRG \textsubscript{CVPR'18} \cite{huang2018weakly} & ResNet-101 &  & 61.4 \\

    AffinityNet \textsubscript{CVPR'18} \cite{ahn2018learning} & Wide ResNet-38 & & 61.7 \\
    
    AffinityNet$^{\dagger}$ \textsubscript{CVPR'18} \cite{ahn2018learning} & ResNet-101 & & 61.9 \\

    SeeNet \textsubscript{NIPS'18} \cite{hou2018self} & ResNet-101 & \checkmark & 63.1 \\

    Zeng \textit{et al} \textsubscript{ICCV'19} \cite{zeng2019joint} &  DenseNet-169 &  & 63.3 \\

    BDSSW \textsubscript{ECCV'18} \cite{fan2018associating} & ResNet-101 & & 63.6  \\

    
    OAA \textsubscript{ICCV'19} \cite{jiang2019integral} & ResNet-101 & \checkmark & 63.9 \\

    CIAN \textsubscript{CVPR'19} \cite{fan2018cian} & ResNet-101 &  & 64.1 \\

    FickleNet \textsubscript{CVPR'19} \cite{lee2019ficklenet} & ResNet-101 & \checkmark & 64.9 \\
    
    SSDD \textsubscript{ICCV'19} \cite{shimoda2019self} & Wide ResNet-38 &  & 64.9 \\
    
    Ours & ResNet101 & \checkmark & \textbf{65.6} \\
    
\bottomrule
\label{table: compare_sota}
\end{tabular}
\vspace{-5mm}
\end{table}

\begin{table*}[!t]
	\caption{Semantic segmentation performance on the PASCAL VOC 2012 validation set. The bottom group contains results with CRF refinement, while the top group is without CRF.
	The best three results are in \textcolor{red}{red}, \textcolor{PineGreen}{green} and \textcolor{blue}{blue}, respectively.}
	
	\vspace{3mm}
	\footnotesize
	\centering
	\renewcommand{\arraystretch}{1.3}
	\setlength{\tabcolsep}{1pt}
	
	\scriptsize
	\begin{tabular}{lccccccccccccccccccccc|c}
		\toprule
		
		
		Method & \rotatebox{90}{bkg} & \rotatebox{90}{aero} & \rotatebox{90}{bike} & \rotatebox{90}{bird} & \rotatebox{90}{boat} & \rotatebox{90}{bottle} & \rotatebox{90}{bus} & \rotatebox{90}{car} & \rotatebox{90}{cat} & \rotatebox{90}{chair} & \rotatebox{90}{cow} & \rotatebox{90}{table} & \rotatebox{90}{dog} & \rotatebox{90}{horse} & \rotatebox{90}{motor} & \rotatebox{90}{person} & \rotatebox{90}{plant} & \rotatebox{90}{sheep} & \rotatebox{90}{sofa} & \rotatebox{90}{train} & \rotatebox{90}{tv} & mIoU \\
		\midrule

		AffinityNet  \cite{ahn2018learning} & 88.2 & 68.2 & 30.6 & \textcolor{PineGreen}{81.1} & 49.6 & 61.0 & 77.8 & 66.1 & 75.1 & 29.0 & 66.0 & 40.2 & 80.4 & 62.0 & \textcolor{blue}{70.4} & \textcolor{red}{73.7} & \textcolor{blue}{42.5} & 70.7 & 42.6 & \textcolor{blue}{68.1} & 51.6 & 61.7 \\
		
		Ours (w/o CRF) & 87.6 & 54.5 & 30.7 & 73.0 & 46.5 & \textcolor{PineGreen}{72.0} & \textcolor{PineGreen}{86.5} & \textcolor{PineGreen}{74.8} & \textcolor{PineGreen}{87.6} & \textcolor{blue}{31.3} & \textcolor{blue}{80.8} & \textcolor{PineGreen}{50.3} & \textcolor{blue}{82.6} & 74.5 & 67.2 & 68.7 & 39.6 & 79.2 & \textcolor{PineGreen}{44.8} & 64.9 & 51.1 & \textcolor{blue}{64.2} \\
		
		\midrule
		
		MCOF \cite{wang2018weakly}  & 87.0 &  \textcolor{red}{78.4} & 29.4 & 68.0 & 44.0 & 67.3 & 80.3 & 74.1 & 82.2 & 21.1 & 70.7 & 28.2 & 73.2 & 71.5 & 67.2 & 53.0 & \textcolor{PineGreen}{47.7} & 74.5 & 32.4 & \textcolor{PineGreen}{71.0} & 45.8 & 60.3 \\
		
		Zeng et al. \cite{zeng2019joint} &  \textcolor{red}{90.0} & \textcolor{PineGreen}{77.4} &  \textcolor{red}{37.5} & \textcolor{blue}{80.7} &  \textcolor{red}{61.6} & 67.9 & 81.8 & 69.0 & 83.7 & 13.6 & 79.4 & 23.3 & 78.0 & \textcolor{blue}{75.3} &  \textcolor{PineGreen}{71.4} & 68.1 & 35.2 & 78.2 & 32.5 & \textcolor{red}{75.5} & 48.0 & 63.3\\ 
		
		FickleNet \cite{lee2019ficklenet} & 
		\textcolor{PineGreen}{89.5} & \textcolor{blue}{76.6} & \textcolor{PineGreen}{32.6} & 74.6 & \textcolor{blue}{51.5} & \textcolor{blue}{71.1} & \textcolor{blue}{83.4} & \textcolor{blue}{74.4} & 83.6 & 24.1 & 73.4 & 47.4 & 78.2 & 74.0 & 68.8 & \textcolor{PineGreen}{73.2} &  \textcolor{red}{47.8} &  \textcolor{blue}{79.9} & 37.0 & 57.3 & \textcolor{red}{64.6} & \textcolor{PineGreen}{64.9} \\

		SSDD \cite{shimoda2019self} & \textcolor{blue}{89.0} & 62.5 & 28.9 & \textcolor{red}{83.7} & \textcolor{PineGreen}{52.9} & 59.5 & 77.6 & 73.7 & \textcolor{blue}{87.0} & \textcolor{red}{34.0} & \textcolor{red}{83.7} & \textcolor{blue}{47.6} & \textcolor{PineGreen}{84.1} & \textcolor{PineGreen}{77.0} & \textcolor{red}{73.9} & 69.6 & 29.8 & \textcolor{red}{84.0} & \textcolor{blue}{43.2} & 68.0 & \textcolor{PineGreen}{53.4} & \textcolor{PineGreen}{64.9} \\

		Ours (w/ CRF) & 88.4 & 57.0 & \textcolor{blue}{31.2} & 75.2 & 47.8 & \textcolor{red}{72.4} & \textcolor{red}{87.2} & \textcolor{red}{76.0} & \textcolor{red}{89.2} & \textcolor{PineGreen}{32.7} & \textcolor{PineGreen}{83.1} & \textcolor{red}{51.1} & \textcolor{red}{85.4} & \textcolor{red}{77.3} & 68.4 &  \textcolor{blue}{70.2} & 40.0 & \textcolor{PineGreen}{81.5} & \textcolor{red}{46.2} & 65.4 & \textcolor{blue}{51.8} & \textcolor{red}{65.6} \\
		
		\bottomrule
	\label{table: compare_sota_detail}
	\end{tabular}
	\vspace{-5mm}
\end{table*}

\vspace{-3mm}
\section{Conclusions}

In this paper, we propose the Mixup-CAM framework to improve the localization of object response maps, as an initial step towards weakly-supervised semantic segmentation task using image-level labels.
To this end, we propose to integrate the mixup data augmentation strategy for calibrating the uncertainty in network prediction. Furthermore, we introduce another two regularization terms as the interplay with the mixup scheme, thereby producing more complete and uniform response maps.
In experimental results, we provide comprehensive analysis of each component in the proposed method and show that our approach achieves state-of-the-art performance against existing algorithms.

\bibliography{egbib}
\end{document}